\documentclass{llncs}
\usepackage[utf8]{inputenc}
\usepackage{graphicx}
\usepackage{float}
\usepackage{gensymb}
\usepackage{comment}
\begin{document}
\title{Terabyte-scale Deep Multiple Instance Learning for Classification and Localization in Pathology}
\author{Gabriele Campanella\inst{1,2} \and
Vitor Werneck Krauss Silva\inst{1} \and
Thomas J. Fuchs\inst{1,2}
}
\authorrunning{G. Campanella et al.}
\institute{Memorial Sloan Kettering Cancer Center, New York, NY 10065, USA \and
Weill Cornell Medicine, New York, NY 10065, USA}
\maketitle

\begin{abstract}
In the field of computational pathology, the use of decision support systems powered by state-of-the-art deep learning solutions has been hampered by the lack of large labeled datasets. Until recently, studies relied on datasets in the order of few hundreds of slides which are not enough to train a model that can work at scale in the clinic. Here, we have gathered a dataset consisting of 12,160 slides, two orders of magnitude larger than previous datasets in pathology and equivalent to 25 times the pixel count of the entire ImageNet dataset. Given the size of our dataset it is possible for us to train a deep learning model under the Multiple Instance Learning (MIL) assumption where only the overall slide diagnosis is necessary for training, avoiding all the expensive pixel-wise annotations that are usually part of supervised learning approaches. We test our framework on a complex task, that of prostate cancer diagnosis on needle biopsies. We performed a thorough evaluation of the performance of our MIL pipeline under several conditions achieving an AUC of 0.98 on a held-out test set of 1,824 slides. These results open the way for training accurate diagnosis prediction models at scale, laying the foundation for decision support system deployment in the clinic.
\end{abstract}

\section{Introduction}

In recent years there has been a strong push towards the digitization of pathology with the birth of the new field of computational pathology \cite{fuchs_computational_2008,fuchs_computational_2011}. The increasing size of available digital pathology data, coupled with the impressive advances that the fields of computer vision and machine learning have made in recent years, make for the perfect combination to deploy decision support systems in the clinic.

Despite a few success stories, translating the achievements of computer vision to the medical domain is still far from solved. 
The lack of large datasets which are indispensable to learn high capacity classification models has set back the advance of computational pathology. The ``CAMELYON16'' challenge for metastasis detection \cite{ehteshami_bejnordi_diagnostic_2017} contains one of the largest labeled datasets in the field with a total of 400 Whole Slide Images (WSIs).
Such an amount of cases is extremely small compared to the millions of instances present in the ImageNet dataset \cite{deng2009imagenet}. One widely adopted solution to face the scarcity of labeled examples in pathology is to take advantage of the size of each example. Pathology slides scanned at 20x magnification produce image files of several Giga-pixels. About 470 WSIs contain roughly the same number of pixels as the entire ImageNet dataset. By breaking the WSIs into small tiles it is possible to obtain thousands of instances per slide, enough to learn high-capacity models from a few hundred slides. 
Pixel-level annotations for supervised learning are prohibitively expensive and time consuming, especially in pathology. Some efforts along these lines \cite{liu_detecting_2017} have achieved state-of-the-art results on CAMELYON16.
Despite the success on these carefully crafted datasets, the performance of these models hardly transfers to the real life scenario in the clinic because of the huge variance in real-world samples that is not captured by these small datasets.

\begin{figure}
\centering
\includegraphics[width=\textwidth]{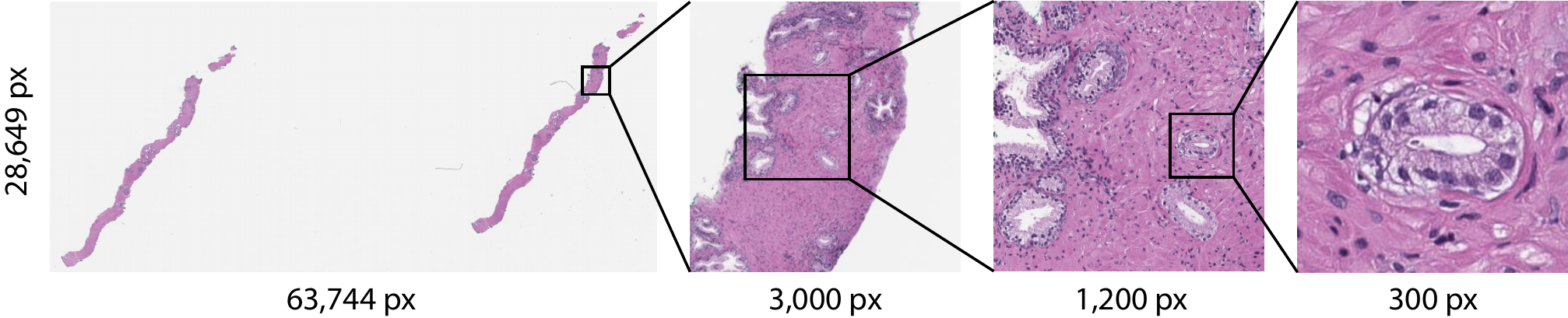}
\caption{\label{fig:prostate}\textbf{Prostate cancer diagnosis is a difficult task.} The diagnosis can be based on very small lesions. In the slide above, only about 6 small tumor glands are present. The right most image shows an example tumor gland. Its relation to the entire slide is put in evidence to reiterate the complexity of the task.}
\end{figure}

In summary, until now it was not possible to train high-capacity models at scale due to the lack of large WSI datasets. Here we gathered a dataset of unprecedented size in the field of computational pathology consisting of over 12,000 slides from prostate needle biopsies, two orders of magnitude larger than most datasets in the field and with roughly the same number of pixels of 25 ImageNet datasets. 
Whole slide prostate cancer classification was chosen as a representative one in computational pathology due to its medical relevance and its computational difficulty. 
Prostate cancer is expected to be the leading source of new cancer cases for men and the second most frequent cause of death behind only the cancers of the lung \cite{siegel_cancer_2016}, and
multiple studies have shown that prostate cancer diagnosis has a high inter- and intra-observer variability \cite{ozdamar_intraobserver_1996,svanholm1985prostatic,gleason_histologic_1992}. 
It is important to note that the classification is frequently based on the presence of very small lesions that can comprise just a fraction of 1\% of the tissue surface. Figure \ref{fig:prostate} depicts the difficulty of the task, where only a few tumor glands concentrated in a small region of the slide determine the diagnosis.

Since the introduction of the Multiple Instance Learning (MIL) framework by \cite{dietterich_solving_1997} in 1997 there have been many efforts from both the theory and application of MIL in the computer vision literature \cite{andrews2002multiple,verma_learning_2012,zhang2006multiple,zhang2002dd}. While the MIL framework is very applicable to the case of WSI diagnosis and despite its success with classic computer vision algorithms, MIL has been applied much less in computational pathology (see Related Work) due to the lack of large WSI datasets. In this work we take advantage of our large prostate needle biopsy dataset and propose a Deep Multiple Instance Learning (MIL) framework where only the whole slide class is needed to train a convolutional neural network capable of classifying digital slides on a large scale. 


\section{\label{sec:related_work}Related work}
This work is related to the extensive literature on computer vision applications to histopathology. In contrast to most of them, here we use weak supervision at the WSI-level instead of strong supervision at the small tile-level. A few other studies have also tackled medical image diagnosis tasks with weak supervision under the MIL assumption:
in \cite{xu_deep_2014} a convolutional neural network is used to classify medical images;
in \cite{xu_weakly_2014} a MIL approach is combined with feature clustering for classification and segmentation tasks.
More recently, \cite{hou_patch-based_2016} showed how learning a fusion model on top of a tile-level classifier trained with MIL can boost performance on two tasks: glioma and non-small-cell lung carcinoma classification. This work is considered the state-of-the-art in terms on weak supervision in hystopathology images. All previous works used small datasets which precludes a proper estimation of the clinical relevance of the models. What sets our efforts apart from others is the scale of our datasets which can be considered clinically relevant.

\section{\label{sec:dataset}Dataset}
\begin{figure}
\centering
\includegraphics[width=0.8\textwidth]{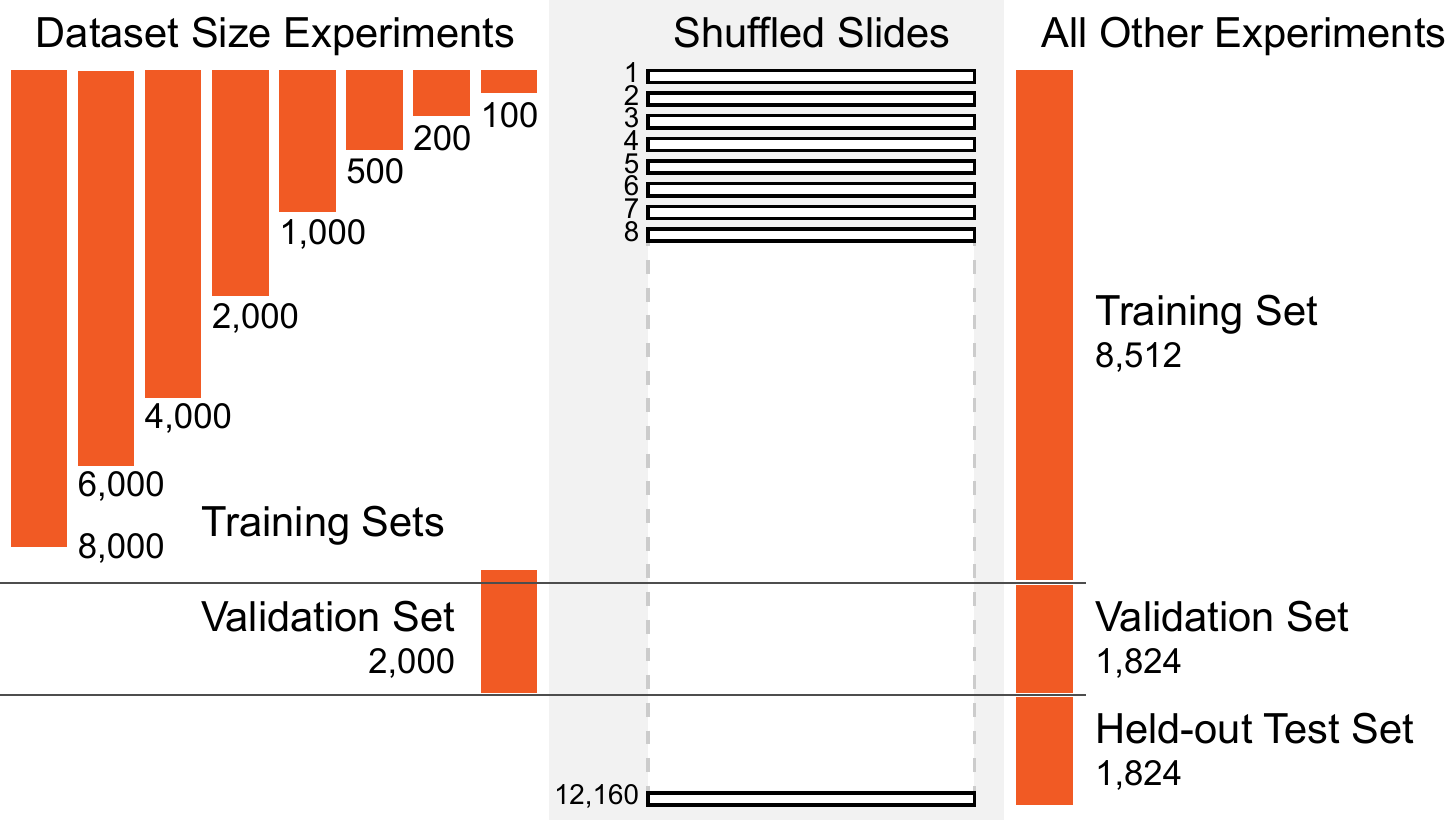}
\caption{\label{fig:dataset}\textbf{Prostate needle biopsy dataset split.} The full dataset was divided into 70-15-15\% splits for training, validation, and test for all experiments except the ones investigating dataset size importance. For those, out of the 85\% training/validation split of the full dataset, training sets of increasing size were generated along with a common validation set.}
\end{figure}

Our dataset consists of 12,160 needle biopsies slides scanned at 20x magnification, of which 2,424 are positive and 9,736 are negative. The diagnosis was retrieved from the original pathology reports in the Laboratory Information System (LIS) at MSKCC. As visualized in Figure \ref{fig:dataset}, the dataset was randomly split in training (70\%), validation (15\%) and testing (15\%). No augmentation was performed during training. For the ``dataset size importance'' experiments, explained further in the Experiments section, a set of slides from the above mentioned training set were drawn to create training sets of different sizes.

\section{\label{sec:method}Methods}
\begin{figure}
\centering
\includegraphics[width=\textwidth]{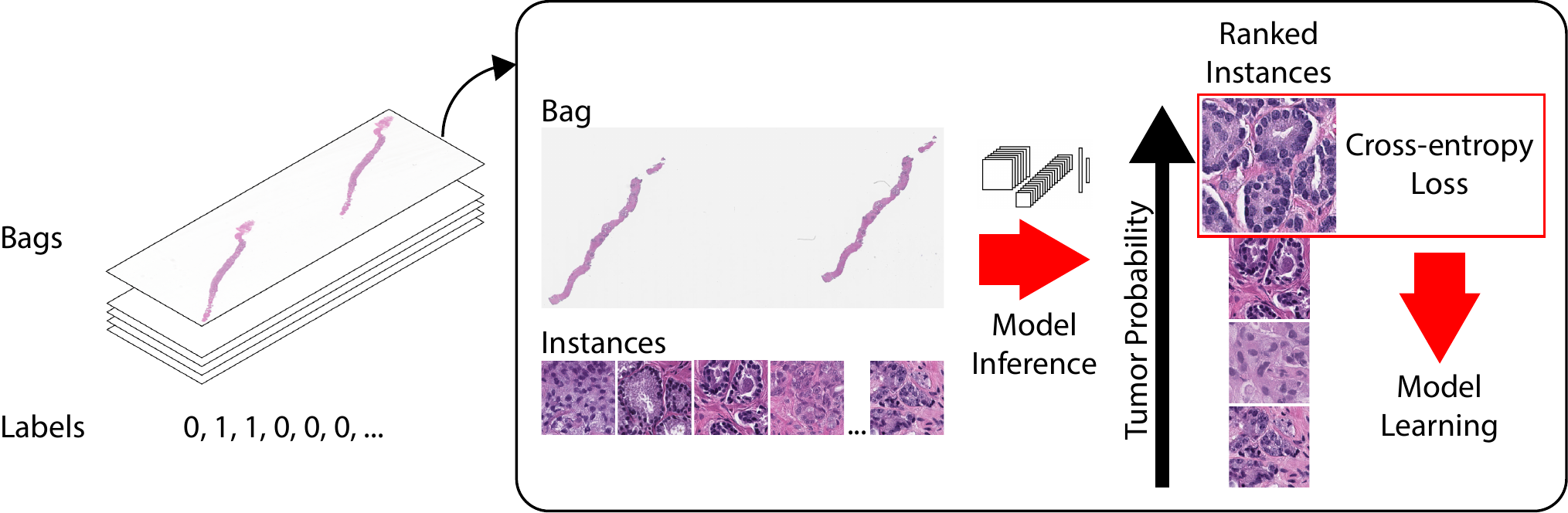}
\caption{\label{fig:pipeline}\textbf{Schematic of the MIL pipeline.} The slide or bag consists of multiple instances. Given the current model, all the instances in the bag are used for inference. They are then ranked according to the probability of being of class positive (tumor probability). The top ranked instance is used for model learning via the standard cross-entropy loss.}
\end{figure}

Classification of a whole digital slide based on a tile-level classifier can be formalized under the classic MIL paradigm when only the slide-level class is known and the classes of each tile in the slide are unknown. Each slide $s_i$ from our slide pool $S=\{ s_i : i=1,2,...,n \}$ can be considered as a bag consisting of a multitude of instances (tiles). For positive bags, there must exist at least one instance that is classified as positive by some classifier. For negative bags instead, all instances must be classified as negative. Given a bag, all instances are exhaustively classified and ranked according to their probability of being positive. If the bag is positive, the top-ranked instance should have a probability of being positive that approaches one, while if it is negative, the probability should approach zero. 
The complete pipeline of our method comprises the following steps: (i) tiling of each slide in the dataset; for each epoch, which consists of an entire pass through the training data, (ii) a complete inference pass through all the data; (iii) intra-slide ranking of instances; (iv) model learning based on the top-1 ranked instance for each slide. A schematic of the method is shown in Figure \ref{fig:pipeline}.

\subsubsection{Slide Tiling:}
We generate the instances for each slide by tiling it on a grid. All the background tiles are efficiently discarded by our algorithm, reducing drastically the amount of computation per slide, since quite a big portion of it is not covered by tissue. Furthermore, tiling can be performed at different magnification levels and with various levels of overlap between adjacent tiles. In this work we investigated three magnification levels (5x, 10x and 20x), with no overlap for 10x and 20x magnification and with 50\% overlap for 5x magnification. On average each slide contains about 100 non overlapping tissue tiles at 5x magnification and 1,000 at 20x magnification. 
Given a tiling strategy we produce our bags $B=\{B_{s_i}:i=1,2,...,n\}$ where $B_{s_i} = \{b_{i,1}, b_{i,2}, ..., b_{i,m}\}$ is the bag for slide $s_i$ containing $m$ total tiles. An example of tiling can be seen in Figure \ref{fig:slide_tiling}.

\subsubsection{Model Training:} 
The model is a function $f_{\theta}$ with current parameters $\theta$ that maps input tiles $b_{i,j}$ to class probabilities for ``negative'' and ``positive'' classes. Given our bags $B$ we obtain a list of vectors $O=\{\overline{o_i}:i=1,2,...,n\}$ one for each slide $s_i$ containing the probabilities of class ``positive'' for each tile $b_{i,j}:j=1,2,...,m$ in $B_{s_i}$. We then obtain the index $k_i$ of the tile within each slide which shows the highest probability of being ``positive'' $k_i = \mathrm{argmax}(\overline{o_i})$. The highest ranking tile in bag $B_{s_i}$ is then $b_{i,k}$. The output of the network $\tilde{y}_i = f_{\theta} (b_{i,k})$ can be compared to $y_i$, the target of slide $s_i$, thorough the cross-entropy loss $l$ as in Equation \ref{eq:loss}.
\begin{equation}
\label{eq:loss}
l = -w_1[y_i\log(\tilde{y}_i)] - w_0[(1-y_i)\log(1-\tilde{y}_i)]
\end{equation}
Given the unbalanced frequency of classes, weights $w_0$ and $w_1$, for negative and positive classes respectively, can be used to give more importance to the underrepresented examples. The final loss is the weighted average of the losses over a mini-batch. Minimization of the loss is achieved via stochastic gradient descent using the Adam optimizer and learning rate 0.0001. We use mini-batches of size 512 for AlexNet, 256 for ResNets and 128 for VGGs. 

\subsubsection{Model Testing:}
At test time all the instances of each slide are fed through the network. Given a threshold (usually 0.5), if at least one instance is positive then the entire slide is called positive; if all the instances are negative then the slide is negative. Accuracy, confusion matrix and ROC curve are calculated to analyze performance.

\section{Experiments}

\subsubsection{Hardware and Software:}
We run all the experiments in our in-house HPC cluster. In particular we took advantage of 7 NVIDIA DGX-1 workstations each containing 8 V100 Volta GPUs. We used OpenSlide \cite{openslide} to access on-the-fly the WSI files and PyTorch \cite{paszke2017automatic} for data loading, building models, and training. Further data manipulation of results was performed in R.

\subsubsection{Weight Tuning:}
Needle biopsy diagnosis is an unbalanced classification task. Our full dataset consists of 19.9\% positive examples and 80.1\% negative ones. To determine whether weighting the classification loss is beneficial, we trained on the full dataset an AlexNet and a Resnet18 networks, both pretrained on ImageNet, with weights for the positive class $w_1$ equal to 0.5, 0.7, 0.9, 0.95 and 0.99. The weights for both classes sum to 1, where $w_1=0.5$ means that both classes are equally weighted. Each experiment was run five times and the best validation balanced error for each run was gathered. Training curves and validation balanced errors are reported in Figure \ref{fig:weight_experiments}. We determined that weights 0.9 and 0.95 gave the best results. For the reminder of the experiments we used $w_1=0.9$.

\subsubsection{Dataset Size Importance:}
\begin{figure}
\centering
\includegraphics[width=0.8\textwidth]{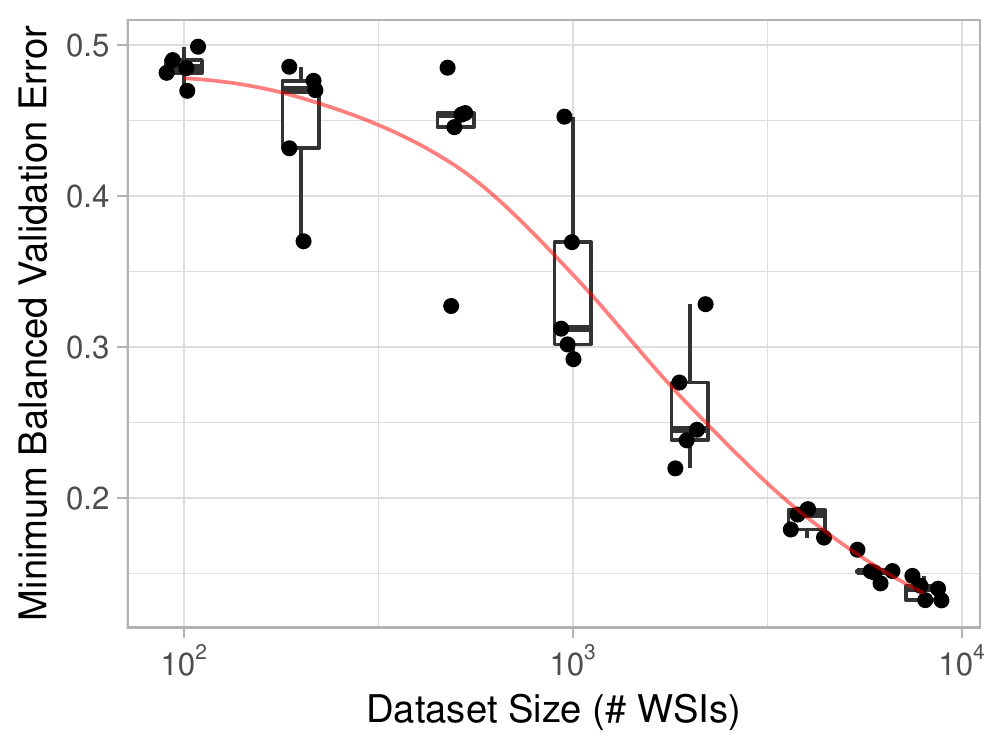}
\caption{\label{fig:size_minerrors}\textbf{Importance of dataset size for MIL classification performance.} Training was performed for datasets of increasing size. The experiment underlies the fact that a large number of slides is necessary for generalization of learning under the MIL setup.}
\end{figure}

In the following set of experiments we determine how dataset size affects performance of a MIL based slide diagnosis task. For these experiments the full dataset was split in a common validation set with 2,000 slides and training sets of different sizes: 100, 200, 500, 1,000, 2,000, 4,000, 6,000. Each bigger training dataset fully contained all previous datasets. For each condition we trained an AlexNet five times and the best balanced errors on the common validation set are shown in Figure \ref{fig:size_minerrors} demonstrating how a MIL based classifier could not have been trained until now due to the lack of a large WSI dataset. Training curves and validation errors are also reported in Figure \ref{fig:size_experiments}.

\subsubsection{Model comparison:}
We tested various standard image classification models pretrained on ImageNet (AlexNet, VGG11-BN, ResNet18, Resnet34) under the MIL setup at 20x magnification. Each experiment was run for up to 60 epochs for at least five times with different random initializations of the classification layers. In terms of balanced error on the validation set, AlexNet performed the worst, followed by the 18-layer ResNet and the 34-layer ResNet. Interestingly, the VGG11 network achieved results similar to those of the ResNet34 on this task. Training and validation results are reported in Figure \ref{fig:model_experiments}.

\paragraph{Test Dataset Performance:}
For each architecture, the best model on the validation dataset was chosen for final testing. Performance was similar with the one on the validation data indicating good generalization. The best models were Resnet34 and VGG11-BN which achieved 0.976 and 0.977 AUC respectively. The ROC curves are shown in Figure \ref{fig:test_roc}a.

\paragraph{Error Analysis:}
A thorough analysis of the error modalities of the VGG11-BN model was performed with the help of an expert pathologist. Of the 1,824 test slides, 55 were false positives (3.7\% false positive rate) and 33 were false negatives (9.4\% false negative rate). The analysis of the false positives found seven cases that were considered highly suspicious for prostate cancer. Six cases were considered ``atypical'', meaning that following-up with staining would have been necessary. Of the remaining false positives, 18 were a mix of known mimickers of prostate cancer: adenosis, atrophy, benign prostatic hyperplasia, and inflammation. The false negative cases were carefully inspected, but in six cases no sign of prostate cancer was found by the pathologist. The rest of the false negative cases were characterized by very low volume of cancer tissue.

\begin{figure}
\centering
\includegraphics[width=\textwidth]{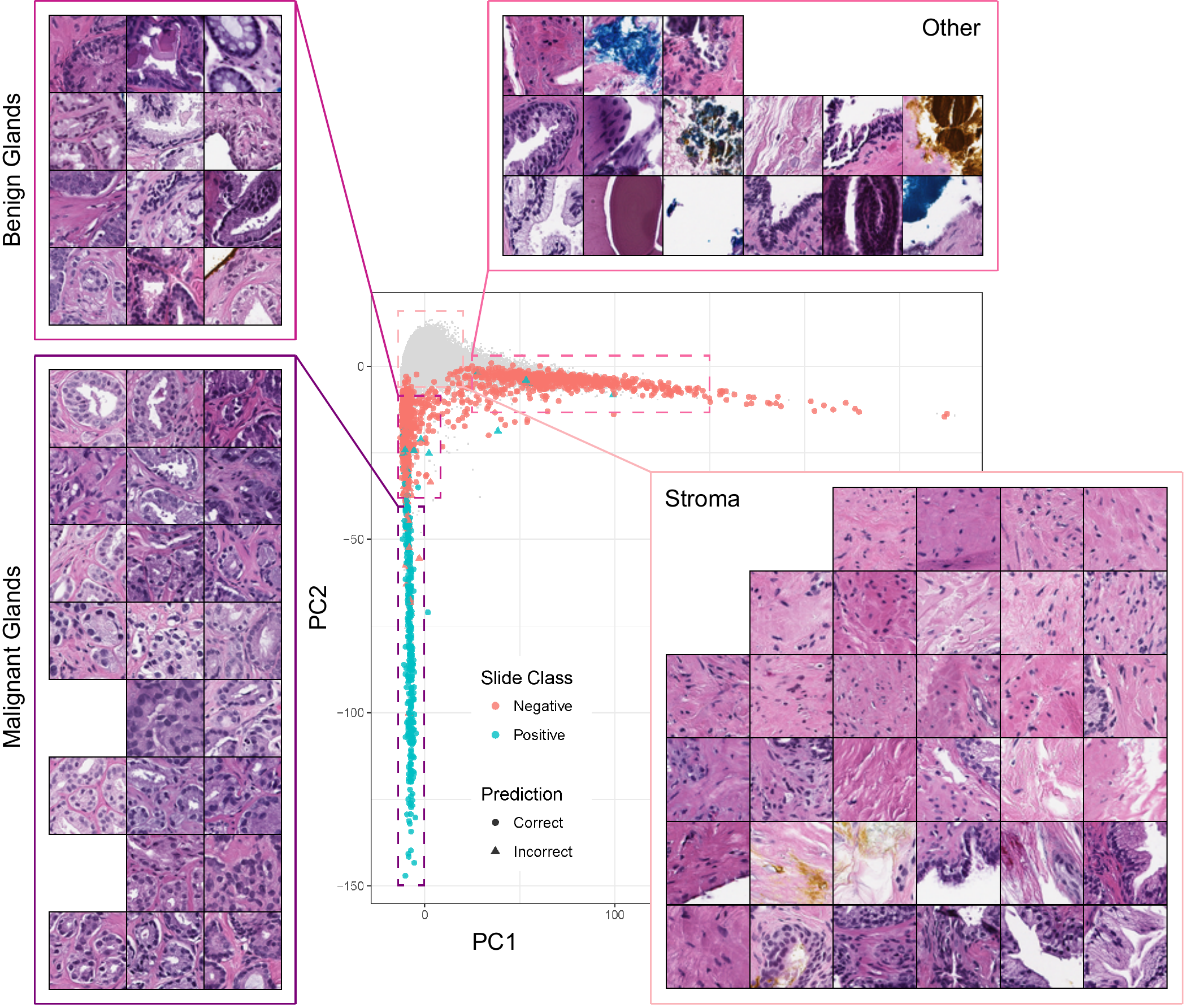}
\caption{\label{fig:pca_embedding}\textbf{Visualization of the feature space with PCA.} A ResNet34 model trained at 20x was used to obtain the feature embedding right before the final classification layer for a random set of tiles in the test set. The embedding was reduced to two dimensions with PCA and plotted. Non-gray points are the top-ranked tiles coming from negative and positive slides. Tiles corresponding to points in the scatter plot are sampled from different regions of the 2D PCA embedding.}
\end{figure}

\paragraph{Feature Embedding Visualization:}
Understanding what features the model uses to classify a tile is an important bottle-neck of current clinical applications of deep learning. One can gain insight by visualizing a projection of the feature space in two dimensions using dimensionality reduction techniques such as PCA. We sampled 50 tiles from each test slide, in addition to its top-ranked tile, and extracted the final feature embedding before the classification layer. As an example, we show the results of the ResNet34 model in Figure \ref{fig:pca_embedding}. From the 2D projection we can see a clear decision boundary between positively and negatively classified tiles. Interestingly, most of the points are clustered at the top left region where we have tiles that are rarely top-ranked in a slide. By observing examples in this region of the PCA space we can determine they are tiles containing stroma. Tiles containing glands extend along the second principal component axis, where there is a clear separation between benign and malignant glands. Other top-ranked tiles in negative slides contain edges and inked regions. The model trained only with the weak MIL assumption was still able to extract features that embed visually and semantically related tiles close to each other.

\subsubsection{Augmentation Experiments:}
We also ran a small experiment with a ResNet34 model to determine whether augmentation of the data with rotations and flips during training could help lower the generalization error. The results, presented in Figure \ref{fig:augmentation_experiments}, showed no indication of a gain in accuracy when using augmentation.

\subsubsection{Magnification comparison:}
\begin{figure}
\centering
\includegraphics[width=\textwidth]{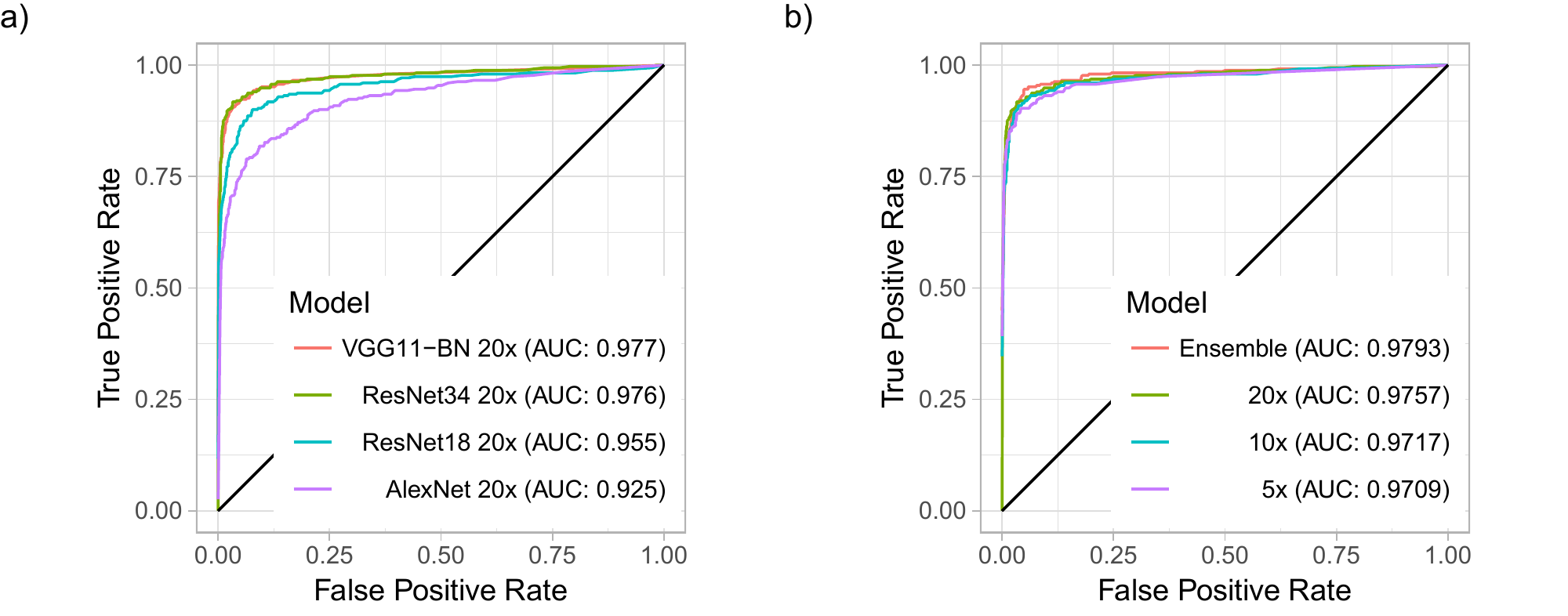}
\caption{\label{fig:test_roc}\textbf{Test set ROC curves.} a) Results of the model comparison experiments on the test set. ResNet34 and VGG11-BN models achieved the best performance. b) Results of the magnification comparison experiments for a ResNet34 model. Lower magnifications lead to worse performance. The max-pooling ensemble model achieves the best result in our experiments.}
\end{figure}

We then trained VGG11-BN and ResNet34 models with tiles generated at 5x and 10x magnifications. Lowering the magnification led consistently to higher error rates across both models. Training curves and validation errors are shown in Figure \ref{fig:scale_experiments}. We also generated ensemble models by averaging or taking the maximum response across different combinations of the three models trained at different magnifications. On the test set these naive multi-scale models outperformed the single-scale models, as can be seen in the ROC curves in Figure \ref{fig:test_roc}b. In particular, max-pooling the response of all the three models resulted in the best results with an AUC of 0.979, a balanced error of 5.8\% and a false negative rate of 4.8\%.

\section{\label{sec:conclusion}Conclusions}
In this study we have analyzed in depth the performance of convolutional neural networks under the MIL assumption for WSI diagnosis. We focused on needle biopsies of the prostate as a complex representative task and obtained the largest dataset in the field with 12,160 WSIs. We demonstrated how it is possible to train high-performing models for WSI diagnosis only using the slide-level diagnosis and no further expert annotation using the standard MIL assumption. We showed that final performance greatly depends on the dataset size. Our best model achieved an AUC of 0.98 and a false negative rate of 4.8\% on a held-out test set consisting of 1,824 slides. We argue that given the current efforts in digitizing the pathology work-flow,  approaches like ours can be extremely effective in building decision support systems that can be effectively deployed in the clinic.

\section*{Acknowledgements}
Dr. Thomas J. Fuchs is a founder, equity owner, and Chief Scientific Officer of Paige.AI.


\bibliographystyle{plain}
\bibliography{biblio}

\newpage
\appendix
\renewcommand\thefigure{\thesection.\arabic{figure}}
\section{Slide Tiling}
\setcounter{figure}{0}
\begin{figure}[H]
\includegraphics[width=\textwidth]{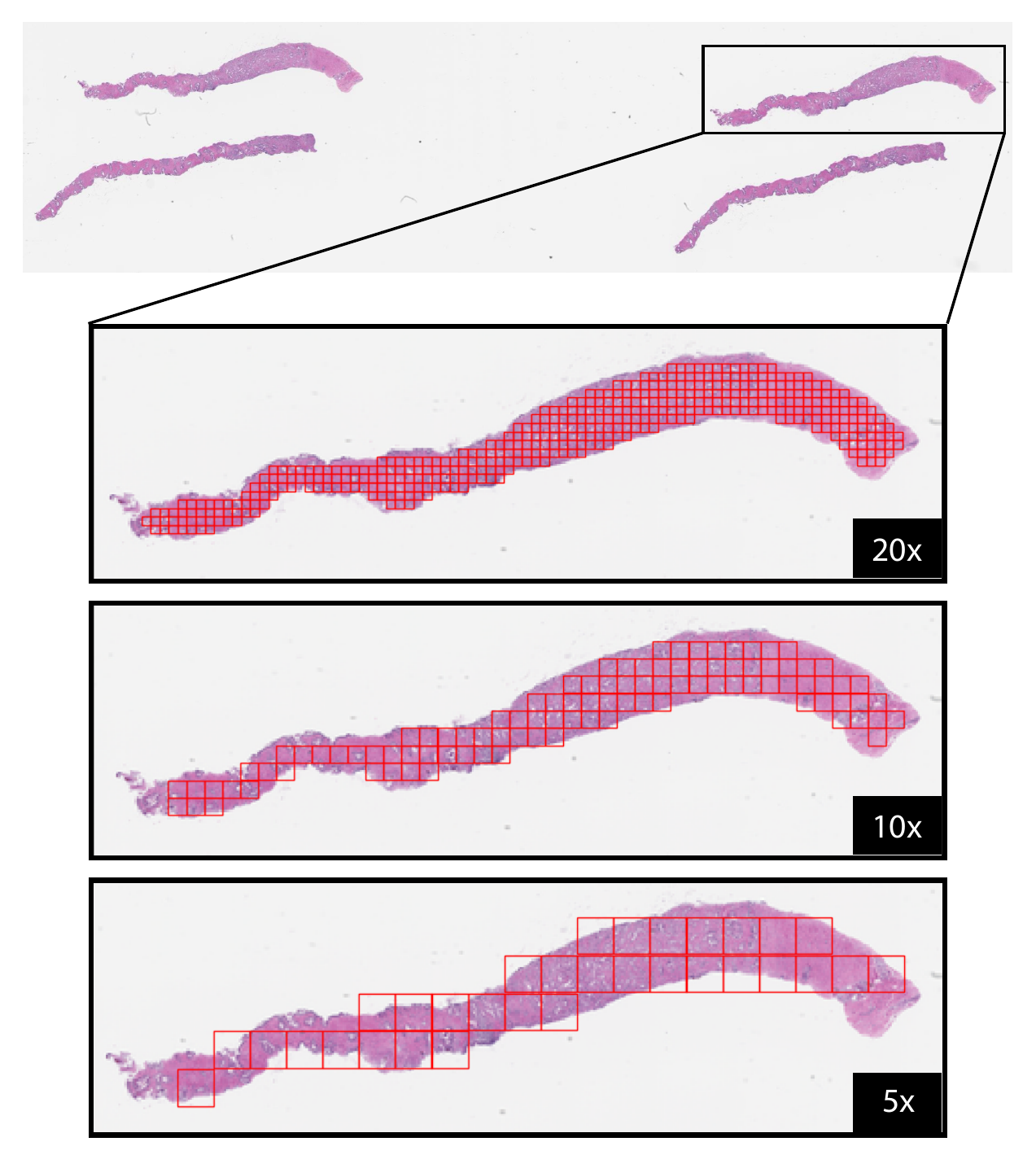}
\caption{\label{fig:slide_tiling}An example of a slide tiled on a grid with no overlap at different magnifications. The slide is the bag and the tiles constitute the instances of the bag. In this work instances at different magnifications are not part of the same bag.}
\end{figure}

\section{Weight Tuning}
\setcounter{figure}{0}
\begin{figure}[H]
\includegraphics[width=\textwidth]{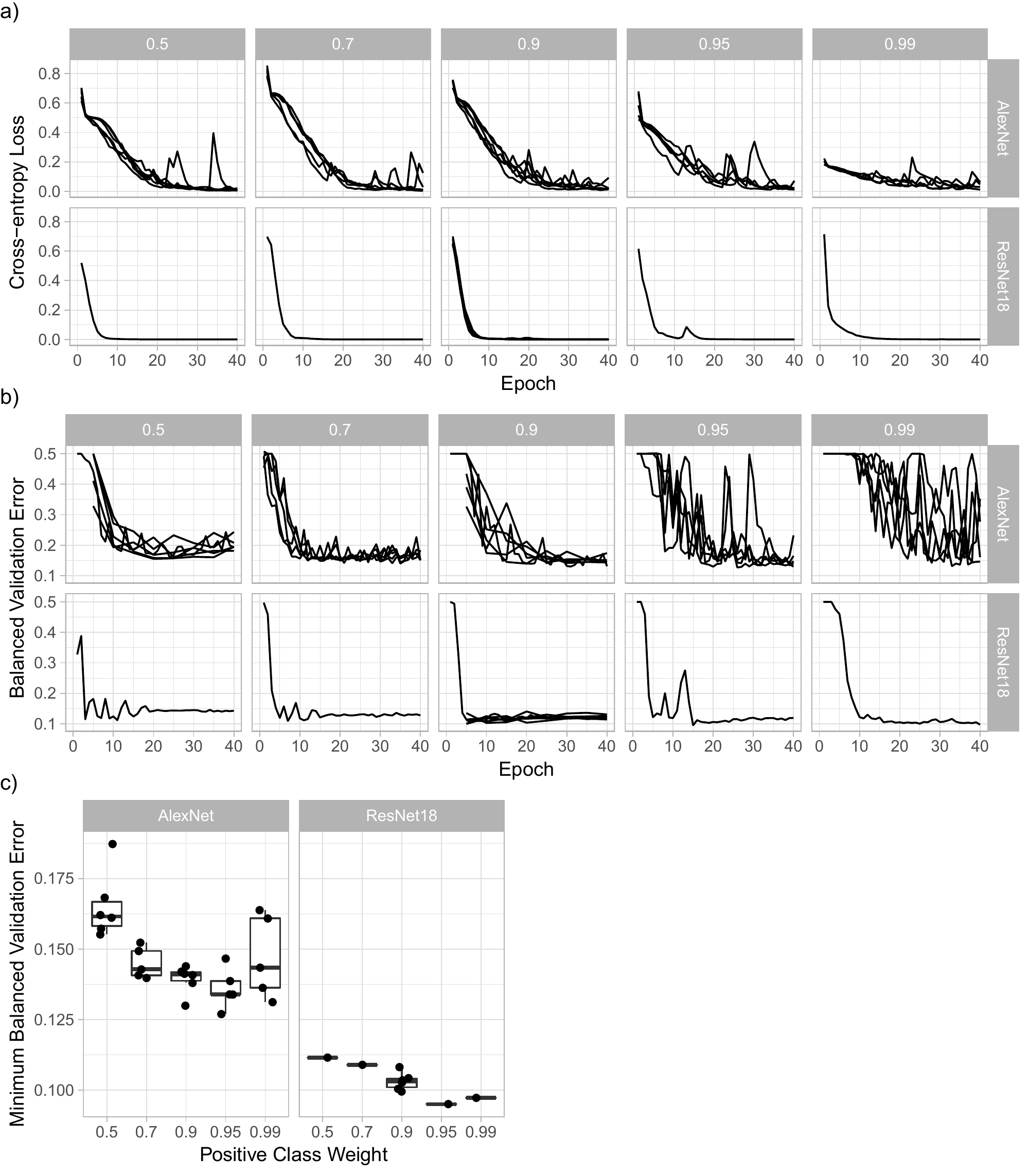}
\caption{\label{fig:weight_experiments}\textbf{Tuning the class weight for the cross-entropy loss}. Different positive class weights were tested: 0.5 (balanced weight), 0.7, 0.9, 0.95, 0.99. Each experiment was run 5 times. a) Training curves with cross-entropy loss. b) Balanced error rates on the validation set. c) For each weight the minimum achieved error was gathered. We chose a value of 0.9 for all subsequent experiments.}
\end{figure}

\section{Dataset Size}
\setcounter{figure}{0}
\begin{figure}[H]
\includegraphics[width=\textwidth]{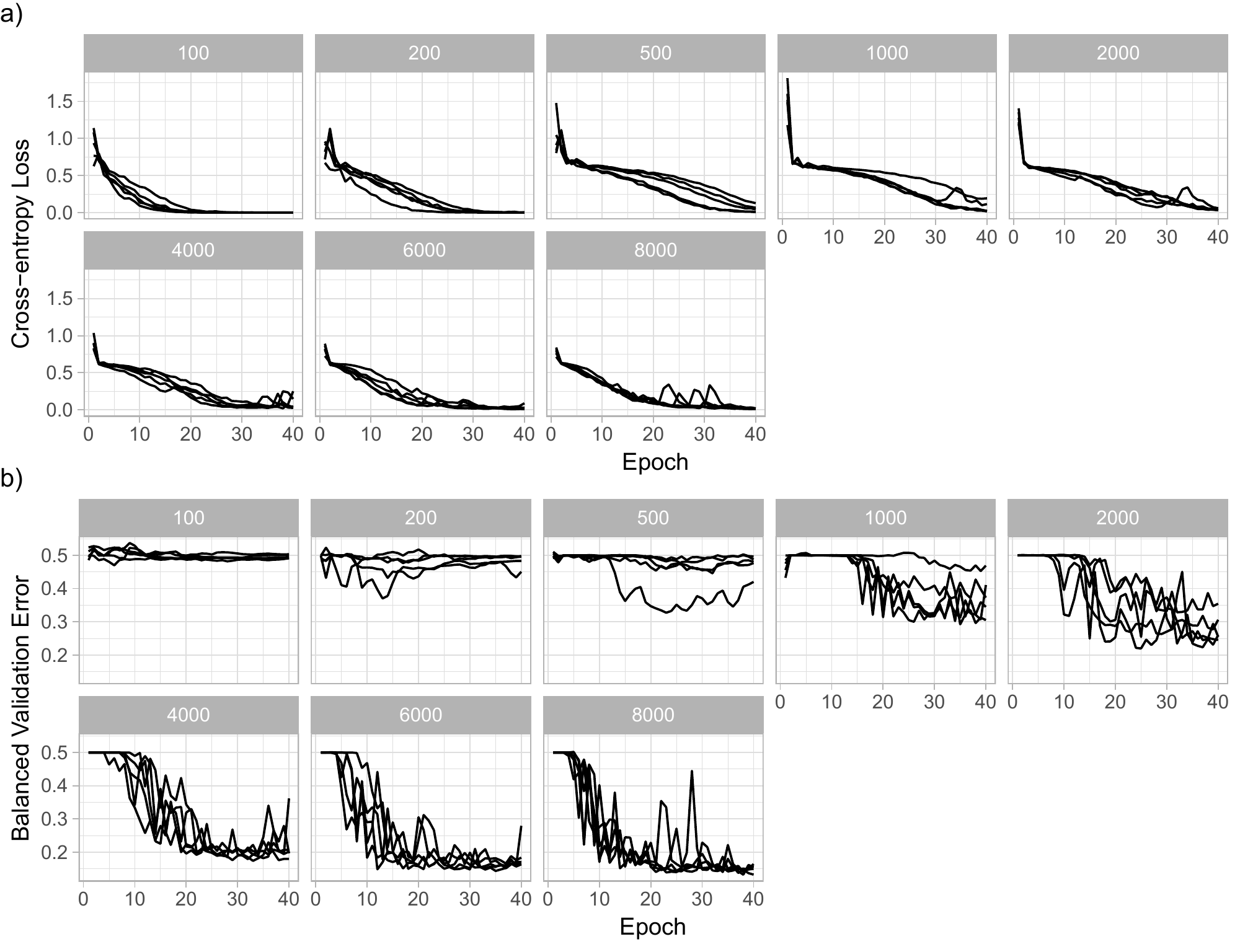}
\caption{\label{fig:size_experiments}\textbf{Training and validation curves for the dataset size experiments}. Good generalization error on the validation set requires a large number of training WSIs. a) Training curves with cross-entropy loss. b) Balanced error rates on the validation set.}
\end{figure}

\section{Model Comparison}
\setcounter{figure}{0}
\begin{figure}[H]
\includegraphics[width=\textwidth]{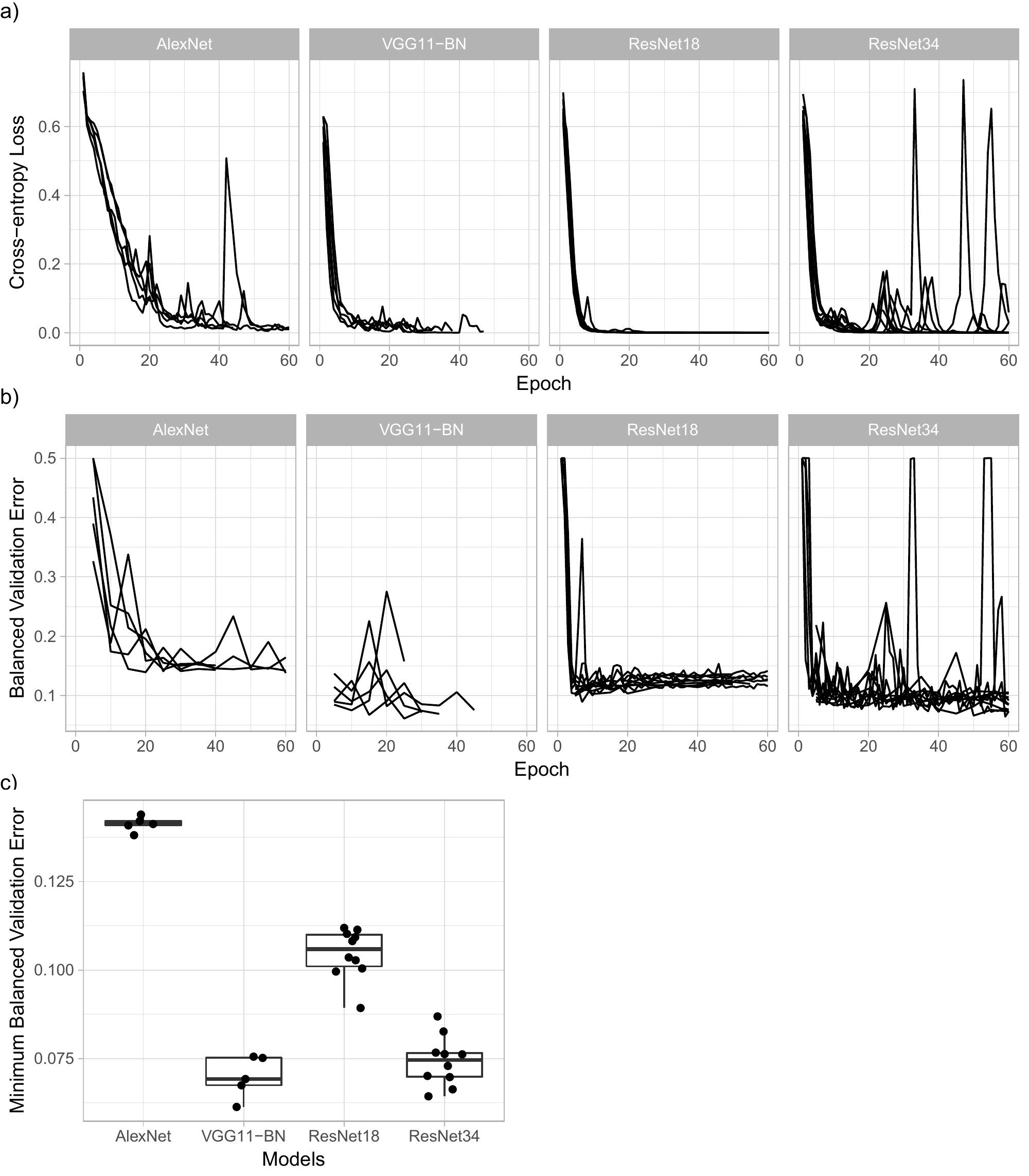}
\caption{\label{fig:model_experiments}\textbf{Comparison of various standard CNN architectures on the MIL task at 20x magnification}. AlexNet, VGG11 with batch-normalization, ResNet18 and ResNet34 models were trained five times each for up to 60 epochs. a) Training curves with cross-entropy loss. b) Balanced error rates on the validation set. c) For each model, the minimum achieved error was gathered.}
\end{figure}

\section{Augmentation Experiments}
\setcounter{figure}{0}
\begin{figure}[H]
\includegraphics[width=\textwidth]{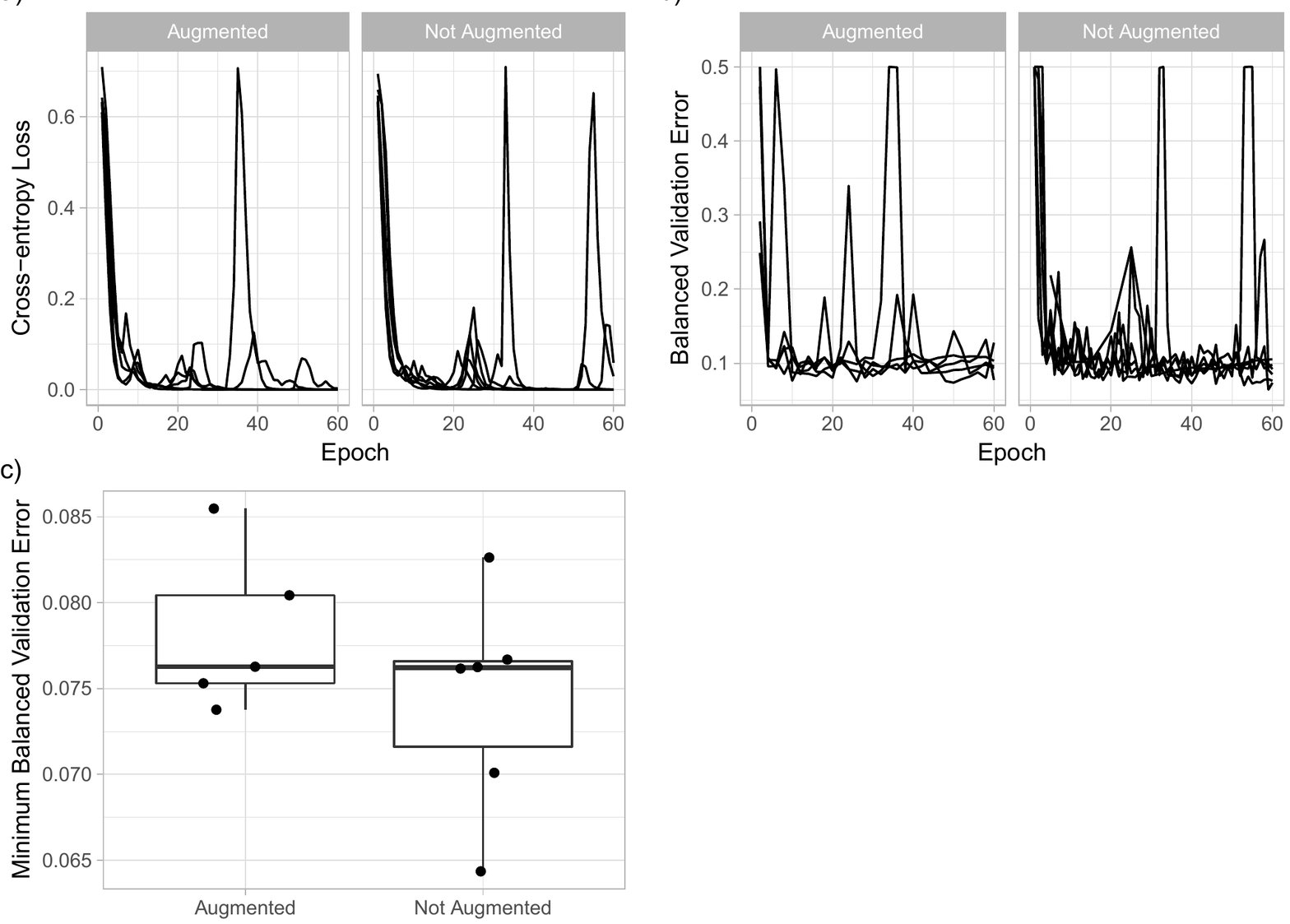}
\caption{\label{fig:augmentation_experiments}\textbf{Comparison of models trained at 20x magnification with and without augmentation}. A ResNet34 model was trained with augmentation consisting of flips and rotations, and without augmentation. Augmentation does not improve generalization. a) Training curves with cross-entropy loss. b) Balanced error rates on the validation set. c) For each model, the minimum achieved error was gathered.}
\end{figure}

\section{Magnification Comparisons}
\setcounter{figure}{0}
\begin{figure}[H]
\includegraphics[width=\textwidth]{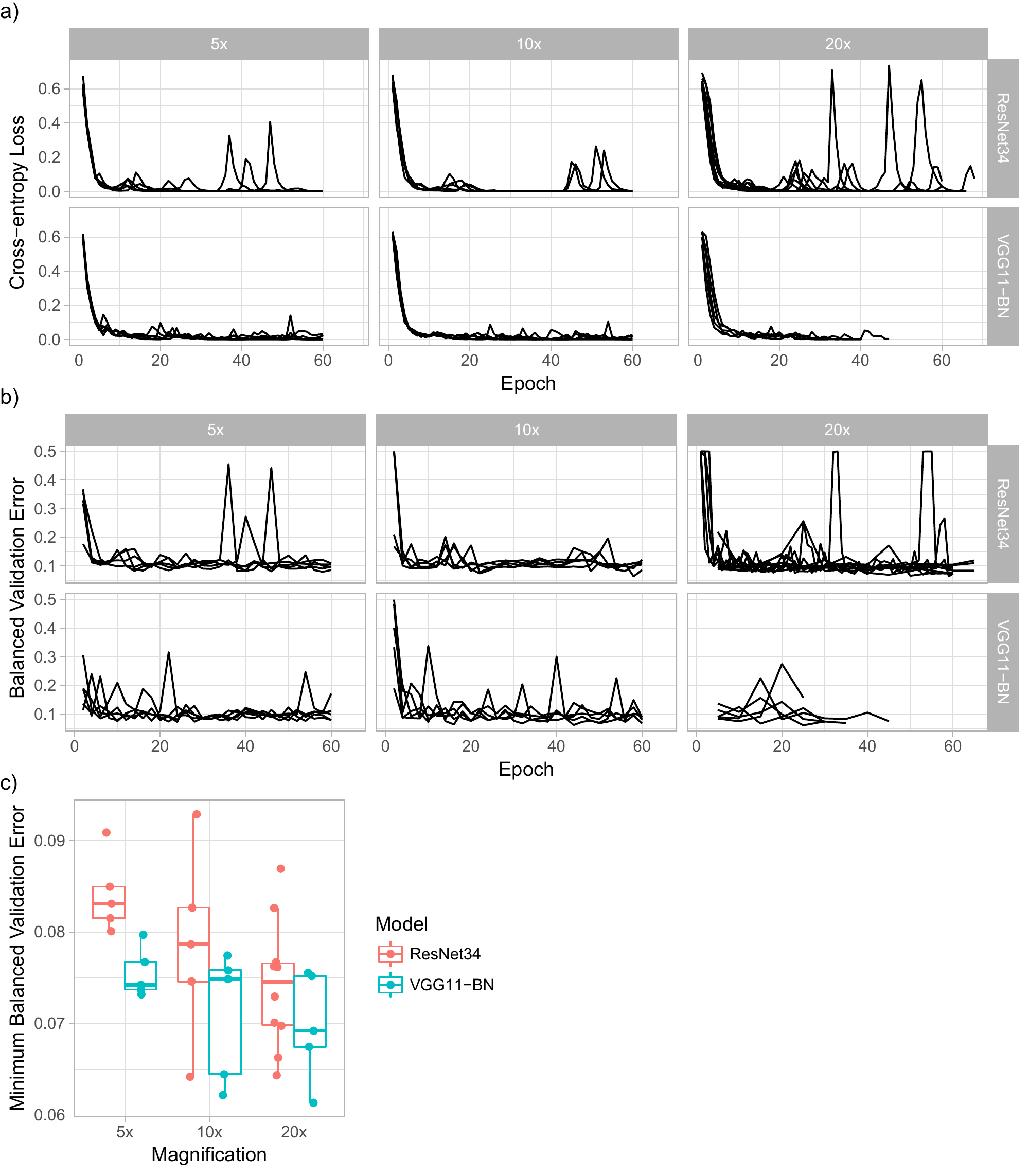}
\caption{\label{fig:scale_experiments}\textbf{Comparison of models trained at different magnifications.}. VGG11-BN and ResNet34 models were trained on tiles extracted at 5x, 10x, and 20x magnifications. Interestingly, performance deteriorates as the magnification of the tiles decreases. a) Training curves with cross-entropy loss. b) Balanced error rates on the validation set. c) For each model, the minimum achieved error was gathered.}
\end{figure}

\end{document}